\documentclass[runningheads]{llncs}
\usepackage{graphicx}
\usepackage{graphicx}
\usepackage{float} 
\usepackage{subfigure} 
\usepackage{booktabs} 

\begin{document}
\title{Evaluating the Effect of Crutch-using on Trunk Muscle Loads}
%
%
\author{Jing Chang\inst{1}\and
Wenrui Wang\inst{2} \and
Damien Chablat\inst{3} \and
Fouad Bennis\inst{4}}
\authorrunning{J. Chang, W. Wang, D. Chablat, F. Bennis}
%
\institute{Tsinghua University, Beijing , 100084, China\\
{\it les\_astres@tsinghua.edu.cn} \and
Ecole Centrale de Nantes, Nantes 44300, France\\
{\it wwr.1122@163.com} \and 
Laboratoire des Sciences du Num\'erique de Nantes, UMR CNRS 6004, Nantes 44300, France \\
{\it damien.chablat@cnrs.fr} \and
\'Ecole Centrale de Nantes, LS2N, UMR CNRS 6004, Nantes 44300, France\\
{\it fouad.bennis@ec-nantes.fr}
}

\maketitle              
\begin{abstract}
As a traditional tool of external assistance, crutches play an important role in society. They have a wide range of applications to help either the elderly and disabled to walk or to treat certain illnesses or for post-operative rehabilitation.
But there are many different types of crutches, including shoulder crutches and elbow crutches. How to choose has become an issue that deserves to be debated. Because while crutches help people walk, they also have an impact on the body. Inappropriate choice of crutches or long-term misuse can lead to problems such as scoliosis.
Previous studies were mainly experimental measurements or the construction of dynamic models to calculate the load on joints with crutches. These studies focus only on the level of the joints, ignoring the role that muscles play in this process.  Although some also take into account the degree of muscle activation, there is still a lack of quantitative analysis.
The traditional dynamic model can be used to calculate the load on each joint. However, due to the activation of the muscle, this situation only causes  part of the load transmitted to the joint, and the work of the chair will compensate the other part of the load. 
Analysis at the muscle level allows a better understanding of the impact of crutches on the body. By comparing the levels of activation of the trunk muscles, it was found that the use of crutches for walking, especially a single crutch, can cause a large difference in the activation of the back muscles on the left and right sides, and this difference will cause muscle degeneration for a long time, leading to scoliosis. In this article taking scoliosis as an example, by analyzing the muscles around the spine, we can better understand the pathology and can better prevent diseases. 
The objective of this article is to analyze normal walking compared to walking with one or two crutches using OpenSim software to obtain the degree of activation of different muscles in order to analyze the impact of crutches on the body.

\end{abstract}
\section{Introduction}
Dating back to ancient Egypt, the crutch was used to overcome gait disorders  for thousands of years \cite{Epstein1972}. Being one of the most traditional and widely used exoskeletons, the crutch is today mainly used for rehabilitation and assistance to the elderly, to help the handicapped, patients and the elderly in necessary activities such as walking and climbing stairs. 

It has been reported that approximately 600,000 Americans use crutches each year \cite{Russell1997}, including people with spinal cord injuries (SCI). It is estimated that IBS affects the quality of life of more than 250,000 Americans and that this number is increasing by 11,000 each year \cite{jackson2004demographic}. Despite the remarkable development of the lower limb exoskeletons, the crutch remains the first choice of SCI patients for mobility assistance. The use of crutches is becoming even more widespread as society ages. People over 65 years of age represent 17.5\% of the EU population in 2011  \cite{eurostat2014population} and this number is estimated to rise to 29.5\% in 2060. The crutch transfers the ground reaction force from the lower limbs to the upper limbs, which largely changes the  kinematic chain of the human body. This transformation of the kinematic chain induces a redistribution of loads between the muscles. It is possible that inappropriate use of crutches could lead to excessive or unbalanced loading on the muscles, which would cause secondary health problems. Therefore, it is necessary to evaluate the effect of crutch use on muscle loads.

Most previous studies on the use of crutches have focused on the upper limbs. Fischer et al. \cite{fischer2014forearm} investigated the pressure on the forearm caused by the use of crutch on 20 healthy adults. The results showed that the maximum pressure on ulnar reached  41 kPa in three different motions, suggesting a high risk of hematoma and pain. Slavens et al. \cite{slavens2011upper} presented an inverse dynamics model that estimated the loads on all upper limb joints when the crutch is applied to the elbow. Experiments on handicapped  children showed that the greatest joint reaction force was in the posterior direction of the wrist, and the greatest joint reaction moment was the flexion moments of the shoulder.  

Studies have also clearly shown how the crutch affects the upper extremities and highlighted its potential risks. In particular, the work of Vankoski et al. \cite{vankoski1997influence} has demonstrated that the loads on lower limb muscle decrease when a crutch is used. However, the effect of the use of crutches on the trunk muscles remains unstudied for no reason. Indeed, the work of Chang et al. \cite{chang2019full} revealed a significant relationship between loads on the arm muscles and loads on the back muscles. In addition, as Requejo et al. \cite{Requejo2005} reported, for users with unbalanced lower limb force generation, arm loads were greater on the opposite side than on the weaker side of the lower limb. In this case, it is very likely that the loads on the trunk muscles are also unbalanced, leading to a high risk of scoliosis.  

The objective of this study is to evaluate the effect of the use of crutches on trunk muscle loads. A numerical model of a human musculoskeletal crutch was built in the OpenSim software to simulate the case of a foot injury. An inverse dynamic analysis was performed on three walking cases, then the activation levels of the trunk muscles were calculated and compared. Particular attention was paid to the bilateral balance of the trunk muscles.
\section{Methods}
\subsection{Walking gaits with crutches}
There are many reasons in real life for people to use crutches, congenital or acquired. If one person has an injury to a foot and the injured foot can still support some of its weight, he may choose to use either a single crutch or a pair of crutches. The corresponding walking patterns are different.

\begin{figure} 
    \centering
    \includegraphics[width=0.5\linewidth]{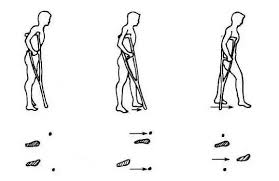}
    \caption{A gait cycle of single-crutch walking \cite{gait}}
    \label{fig_single}
\end{figure}

When walking with only one crutch, the crutch must be located on the non-injured side. In the first phase of a gait cycle, the crutch and the injured foot are stretched, during which the weight of the body is supported by the healthy foot. The second phase starts when the crutch and the injured foot are on the ground and then the centre of gravity begins  to move forward. In this phase, the body weight is distributed between the crutch and both feet. In the finial phase, the centre of gravity is between the crutch and the injured foot. At this time, the latter foot leaves the ground and moves forward. A whole walking cycle ends when the injured foot reaches the ground, as shown in Figure~\ref{fig_single}. 

\begin{figure}
    \centering
    \includegraphics[width=0.75\linewidth]{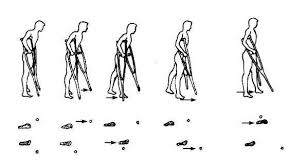}
    \caption{A gait cycle of double-crutch walking \cite{gait}}
    \label{fig_double}
\end{figure}

Double crutch walking begins when the crutch on the healthy side is extended, followed by the injured foot. At the same time, the centre of gravity shifts forward to reach the extended crutch and foot. Then the crutch behind moves forward, followed by the last foot, as shown in  Figure~\ref{fig_double}. 
\subsection{Crutch model and musculoskeletal modeling with OpenSim software}
OpenSim is an open source software for biomechanical modeling, simulation and analysis \cite{seth2018opensim}. Users are enabled to build their own model. In this study, a human- crutch model is built based on the full-chain model developed by the LS2N \cite{chang2019full}. The basic model has 46 bodies, 424 muscles and 37 degrees of freedom, while the double-crutch model has two more bodies and four other degrees of freedom. The constructed model covers the muscles of the kinetic chain of the whole body. 
With OpenSim software, the inverse kinematics as well as the inverse dynamics can be easily evaluated. The calculated joint moments can be distributed between each muscle by an optimization algorithm according to their properties (MVC: maximum voluntary contraction). 
\subsection{Simulation settings}
In this study, three walking cases are examined: normal walking, walking with one crutch and walking with two crutches. 
When a person is injured on one foot, but it still provides some support, a single crutch can be used to assist walking.
However, when this foot is particularly injured and cannot provide this part of support, or when both feet are injured, but can also provide partial support, at this time a single crutch can no longer assist people to walk, choose a double crutch It is necessary to assist walking \cite{shoup1974biomechanics}.
\begin{figure}
    \centering
    \includegraphics[width=0.45\linewidth]{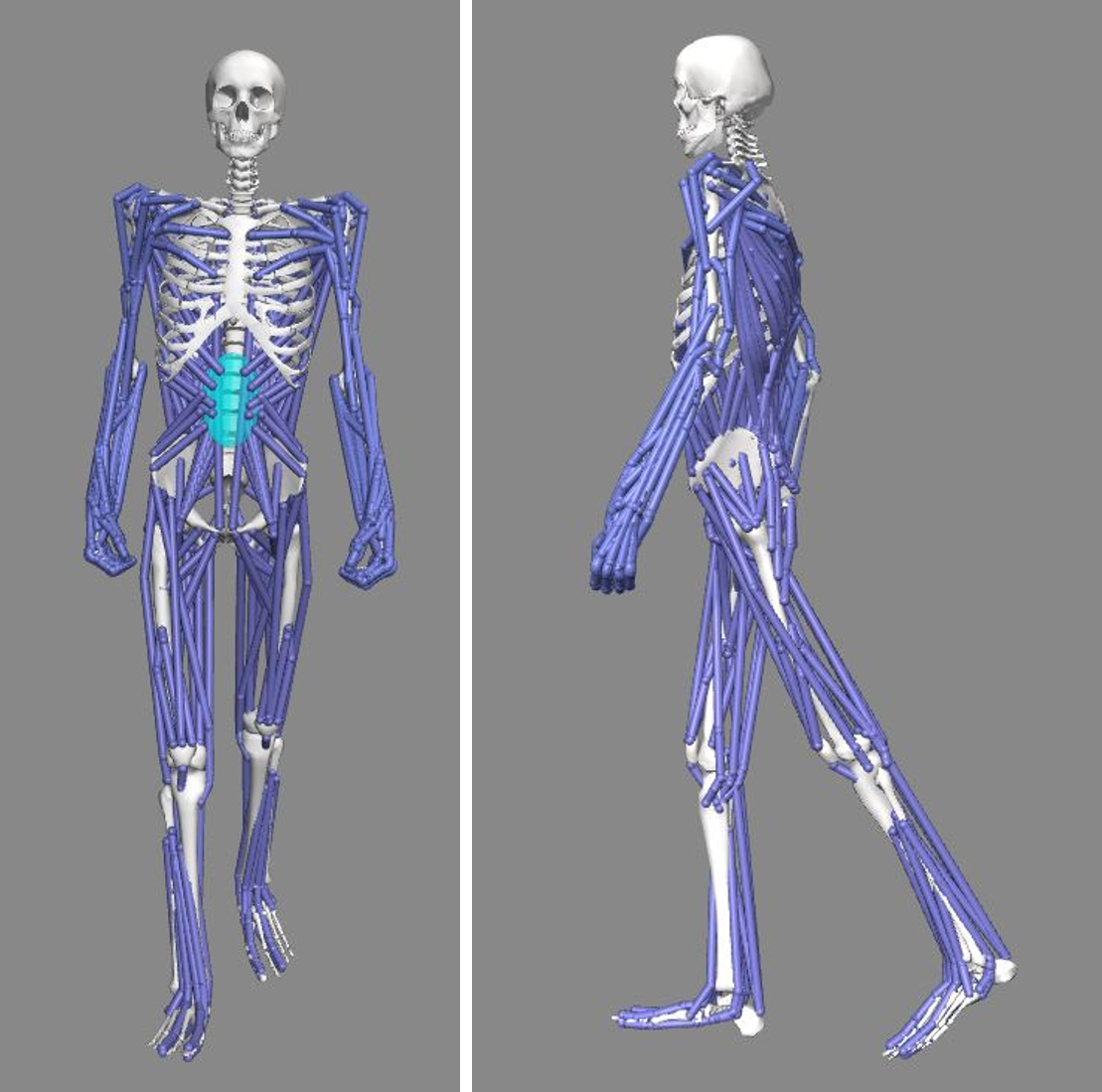}
    \caption{Normal walking}
    \label{health}
\end{figure}
\subsection{Single crutch walking}
Suppose that the experimenter's right foot is injured and can only bear $10\%$ of the body weight \cite{li2001three}. The left hand uses a single crutch to perform a three-point gait. At this time, the gravity is mainly borne by the left foot and the crutch as shown in Figure~\ref{single_cruch}.
\subsection{Double crutches walking}
\begin{figure}[H]
  \begin{minipage}[t]{0.29\linewidth}
    \centering
    \includegraphics[width=0.99\linewidth]{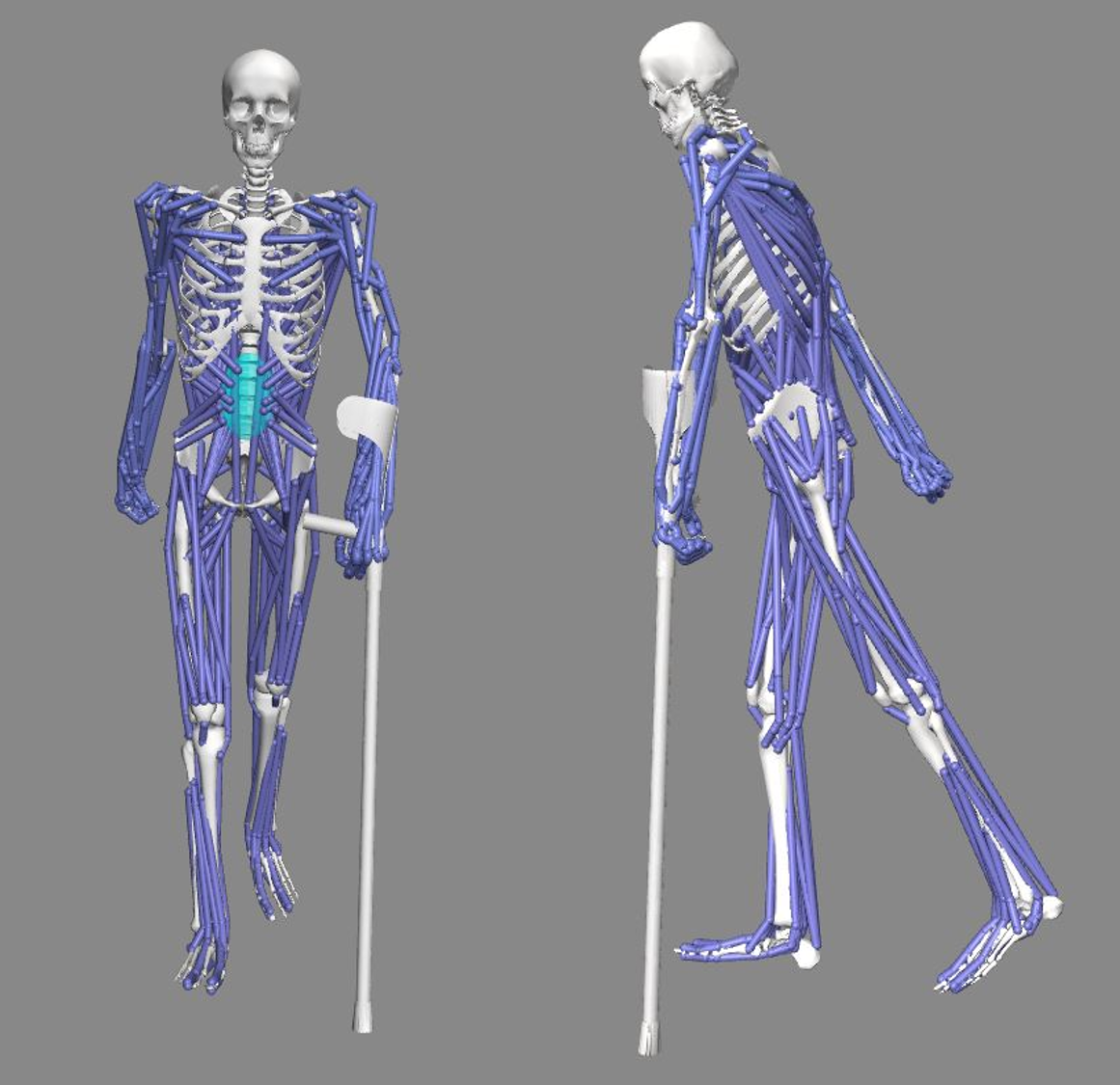}
    \caption{Walking with a single crutch}
    \label{single_cruch}
  \end{minipage}
  \begin{minipage}[t]{0.7\linewidth}
    \centering
    \includegraphics[width=0.95\linewidth]{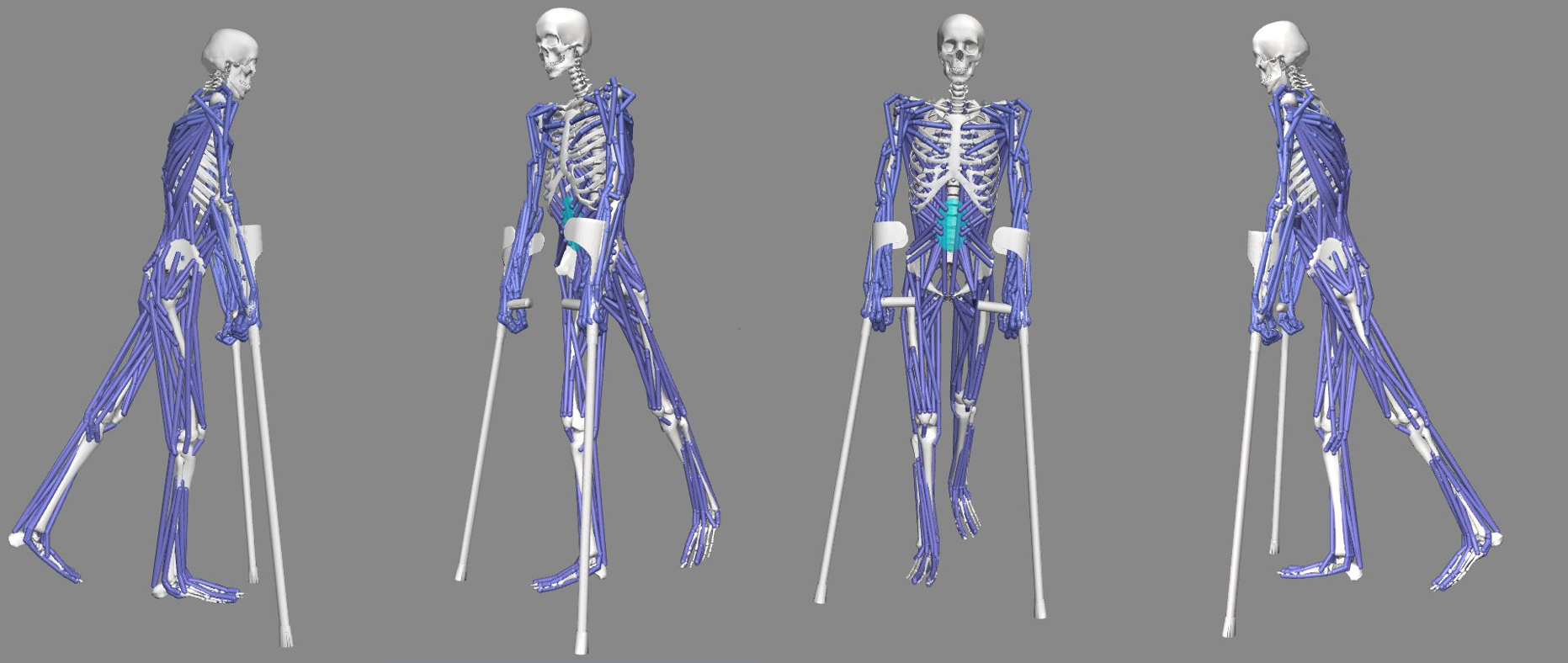}
    \caption{Walking with double crutches}
    \label{double_crutches}
  \end{minipage}
\end{figure}

Suppose the experimenter's right foot is injured and can only support  10\% of the body weight \cite{li2001three}. Double crutches are used for four-point walking. At this time, the right foot and the double crutches have moved forward, and the center of gravity has moved forward, so that the gravity is mainly supported by the right foot and double crutches. The next procedure is to lift the left foot and move close as shown in Figure~\ref{double_crutches}.
\section{Results of the simulations}
We calculated the degree of activation of different muscles through experiments. These muscles are as follows: 
\begin{itemize}
\item Rectus abdominis muscles (2 muscles); 
\item Iliacus muscles (22 muscles); 
\item Abdominal external oblique muscles (12 muscles); 
\item Abdominal internal oblique muscles (12 muscles);
\item Quadratus lumborum muscles (36 muscles);
\item Iliocostalis muscles (24 muscles); 
\item latissimus dorsi muscles (28 muscles);
\item Longissimus muscles (10 muscles). 
\end{itemize}
During normal walking, walking with a single crutch and walking with a double crutches, the average degrees of muscle activation obtained were: 1\%, 31\%, 9\%, respectively. From these results, we can see that using crutches to assist walking weighs more than normal walking, and we can also find that the muscles that use single crutch bears more load than those that use double crutches.

\begin{table}[ht!]
    \centering
        \includegraphics[width=1\linewidth]{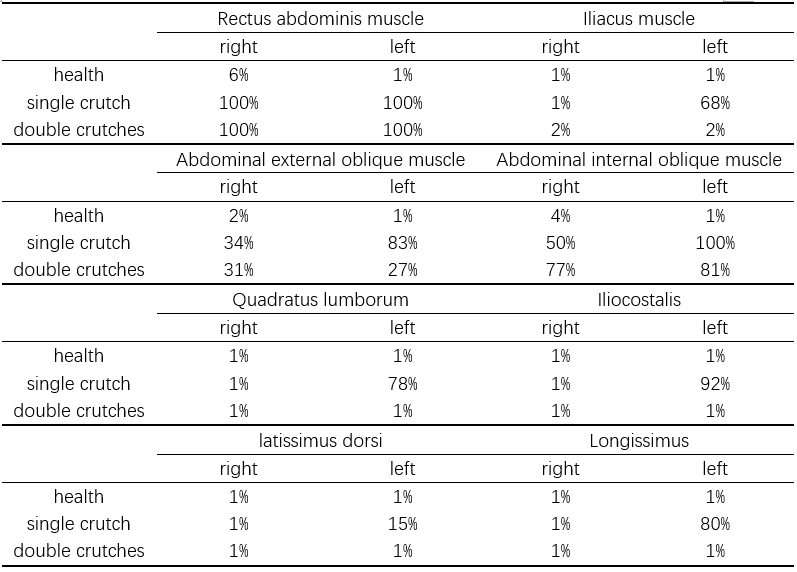}
    \caption{Trunk muscle activation}
    \label{tab:Trunk_muscle_activation}
\end{table}
\begin{figure}
    \centering
    \includegraphics[width=0.25\linewidth]{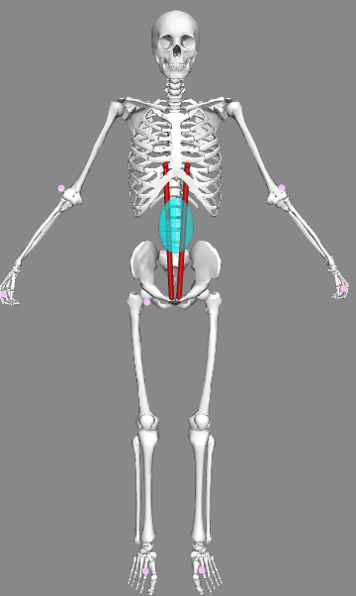}
    \caption{Rectus abdominis muscle}
    \label{fig_rAb}
\end{figure}

The rectus abdominis is a  long flat muscle that connects the sternum with the pubic junction, shown in Figure~\ref{fig_rAb}. It is an important posture muscle, responsible for flexion of the lumbar spine. The activation levels of the left and right abdominal muscles are 6\% and 1\%, respectively, for the normal  walking. When walking with one or two crutches,  the level of activation of the two right abdominal muscles is 100\%. This suggests a high load on the right abdominal muscle when using crutches.

\begin{figure}[H]
  \begin{minipage}[b]{0.49\linewidth}
    \centering
    \includegraphics[width=0.99\linewidth]{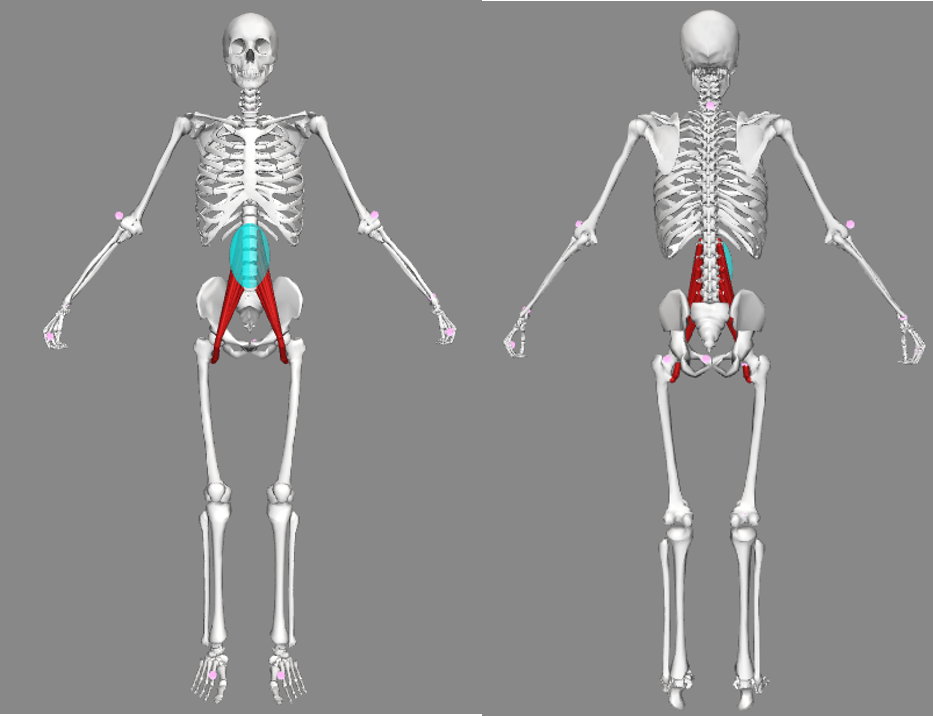}
    \caption{Iliacus muscle activation}
    \label{fig_Ili}
  \end{minipage}
  \begin{minipage}[b]{0.49\linewidth}
    \centering
    \includegraphics[width=0.94\linewidth]{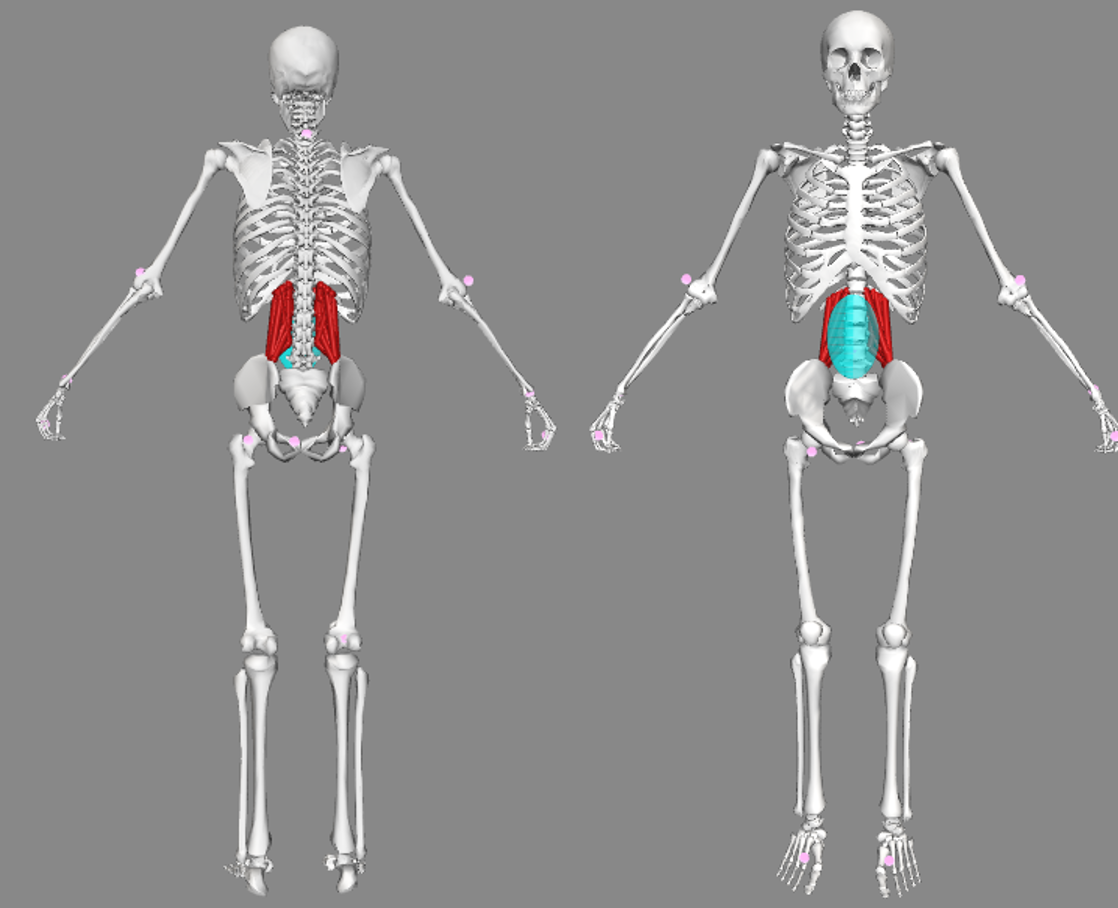}
    \caption{Quadratus lumborum Muscles}
    \label{fig_Quadratus}
  \end{minipage}
\end{figure}

The iliacus is a flat, triangular muscle. It forms the lateral part of the iliopsoas, providing flexion of the thigh and lower limb at the hip joint, shown in Figure~\ref{fig_Ili}. It is important for lifting (flexing) the femur forward. From Table~\ref{tab:Trunk_muscle_activation}, we can see that the  activation levels of the left and right iliacus muscles are 1\% when a person walks normally. But when people walk with only one  crutch, the activation levels of the left and right iliacus muscles  are 68\% and 1\%. Although there is no additional load on the right muscles, they are overloaded by the left muscles. We can see that when people walk with only one crutch, the left iliacus muscles cause a lot of load. 

This results in asymmetrical muscle activation. Prolonged asymmetry leads a risk of scoliosis. When people walk with  double crutches, the activation levels of the left and right iliacus muscles are both 2\%. The muscles symmetry of the double crutches are perfect, and the muscles load is minimal, which is similar to normal walking. As the right foot is injured, the left leg has to support more force than usual when walking with a single crutch. However, because the right foot is injured, the right leg does not play an important role in the gait process. The muscles do not support any additional load. Long-term muscle asymmetry can also lead to a deformation of the spine connected to these muscles, causing scoliosis.

The quadratus lumborum muscle is a paired muscle of the left and right posterior abdominal wall. It is the deepest abdominal muscle and commonly referred to as a back muscle, shown in Figure~\ref{fig_Quadratus}.  It can first perform a lateral flexion of the spine and an ipsilateral contraction. Second, it can extend the lumbar spine and contract bilaterally. Thirdly, it can perform vertical stabilization of the pelvis, lumbar spine and lumbosacral junction, preventing scoliosis. Fourthly, it can tilt the pelvis forward, it is the contralateral lateral pelvic rotation. 

From Table~\ref{tab:Trunk_muscle_activation}, we can see that the activation levels of the left and right quadratus lumborum muscles are both 1\% when a person walks normally. But when people walk with only one crutch, the activation levels of the left and right quadratus lumborum muscles  are 78\% and 1\%. This situation causes the asymmetric activation of the muscles. Prolonged asymmetry carries the risk of scoliosis. When people walk with a double crutches, the activation levels of the left and right quadratus lumborum  are both 1\%. The muscles symmetry of the double crutches are perfect, and the muscles load is minimal, which are similar to normal walking.

\begin{figure}[H]
  \begin{minipage}[b]{0.49\linewidth}
    \centering
    \includegraphics[width=0.9\linewidth]{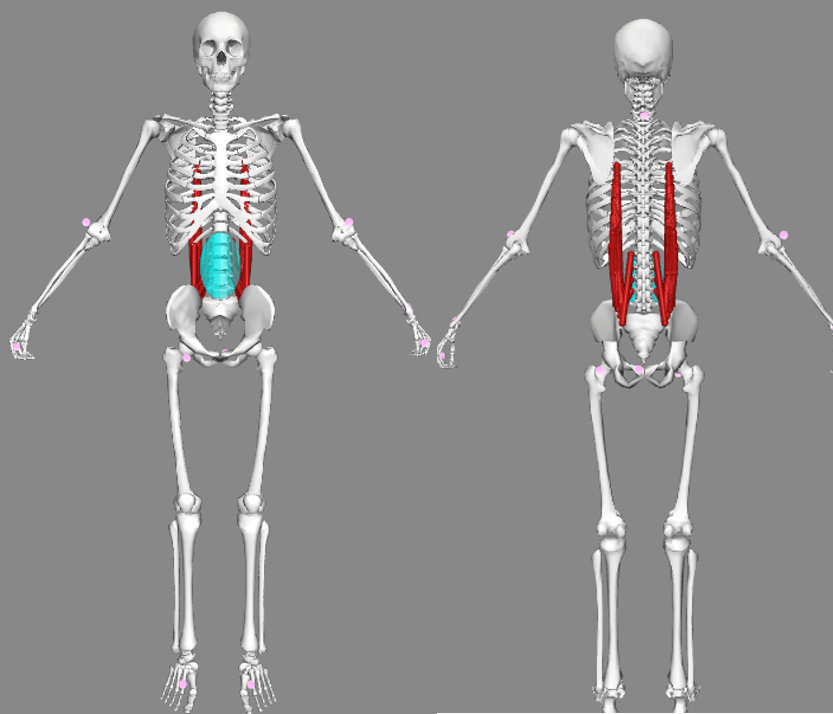}
    \caption{The iliocostalis muscles}
    \label{fig_iliocostalis}
  \end{minipage}
  \begin{minipage}[b]{0.49\linewidth}
    \centering
    \includegraphics[width=0.95\linewidth]{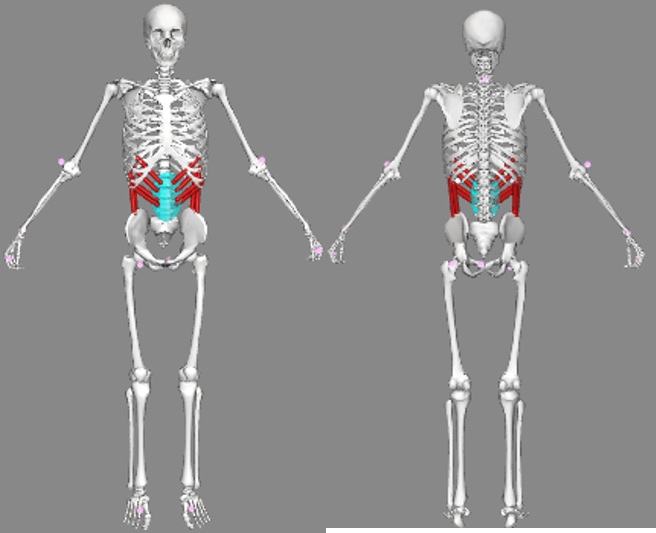}
    \caption{The external oblique muscles.}
    \label{fig_Abdominal external oblique muscle}
  \end{minipage}
\end{figure}

The iliocostalis is the muscle immediately lateral to the longissimus that is the nearest to the furrow that separates the epaxial muscles from the hypaxial, shown in Figure~\ref{fig_iliocostalis}. From Table~\ref{tab:Trunk_muscle_activation}, we can see that the actication levels of the  iliocostalis muscles are 92\% and 1\%. Although there is no additional load on the right muscles, it is overloaded by the left muscles. It can be seen that when people walk with one crutch, the left iliocostalis muscles are more activated. This leads the asymmetric activation of the muscles. Prolonged asymmetry carries the risk of scoliosis. When people walk with double crutches, the activation level of the left and right iliocostalis muscles is 1\%. The muscles symmetry of the double crutches is perfect, and the muscles load are minimal, which are similar to normal walking.

The abdominal external oblique muscle is the largest and outermost of the three flat abdominal muscles of the lateral anterior abdomen. The external oblique is situated on the lateral and anterior parts of the abdomen. It is broad, thin, and irregularly quadrilateral, its muscular portion occupying the side, its aponeurosis the anterior wall of the abdomen. It also performs ipsilateral side-bending and contralateral rotation. So the right external oblique would side bend to the right and rotate to the left. The internal oblique muscle functions similarly except it rotates ipsilaterally, shown in Figure~\ref{fig_Abdominal external oblique muscle}. From Table~\ref{tab:Trunk_muscle_activation}, we can see that the activation levels of the left and right abdominal external oblique muscles are 1\%  when a person is walking normally. But when people walk with a single crutch, the activation levels of the left and right abdominal external oblique muscles are 83\% and 34\%. It can be seen that when people walk with a single crutch, the left abdominal external oblique muscles cause a lot of loads. This situation cause the asymmetric activation of the muscles. 
And the load on the left and right muscles is much higher than normal walking. When people walk with a double crutches, the activation levels of the left and right abdominal external oblique muscles are 27\% and 31\%. Although the muscles load are higher than during normal walking. They are more symmetrical than when people walk with a single crutch. This muscles are responsible for torso rotation. From this, we can see that when the human body rotates left to right under normal conditions, the force used is the same, and it is easy. When the right foot is injured, the rotation of the human body becomes much more difficult, especially in the case of a single crutch. Due to the injury of the right foot, the human body's torso is not flexible to rotate to the right, so this part of the muscles need to provide a lot of energy to help the torso rotate.

\begin{figure}[H]
  \begin{minipage}[b]{0.47\linewidth}
    \centering
    \includegraphics[width=0.99\linewidth]{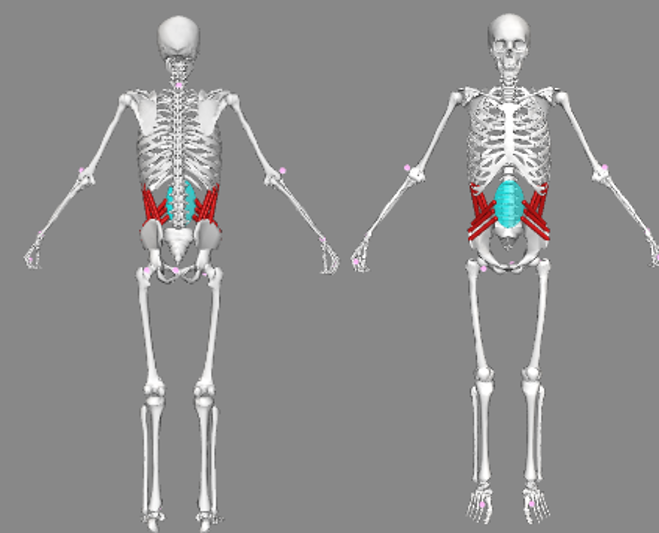}
    \caption{Internal oblique muscles} 
    \label{fig_Abdominal internal oblique muscle}
  \end{minipage}
  \begin{minipage}[b]{0.47\linewidth}
    \centering
    \includegraphics[width=0.92\linewidth]{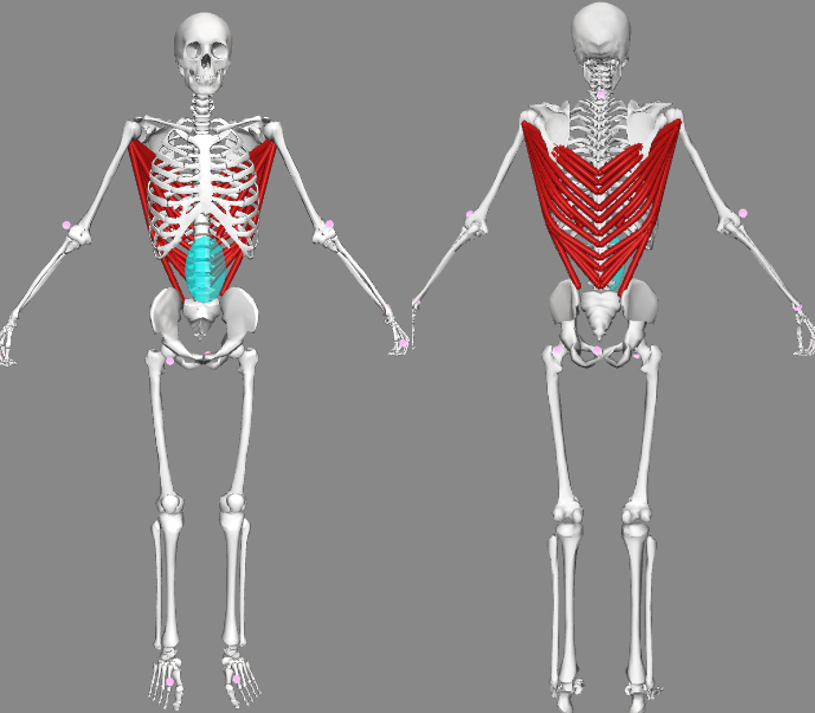}
    \caption{Latissimus dorsi muscle group}
    \label{fig_latissimus dorsi}
  \end{minipage}
\end{figure}

The abdominal internal oblique muscle is an abdominal muscle in the abdominal wall that lies below the external oblique muscle and just above the transverse abdominal muscle, shown in Figure~\ref{fig_Abdominal internal oblique muscle}. Its fibers run perpendicular to the external oblique muscle, and the lateral half of the inguinal ligament. Its contraction causes ipsilateral rotation and side-bending. It acts with the external oblique muscle of the opposite side to achieve this torsional movement of the trunk. From Table~\ref{tab:Trunk_muscle_activation}, we can see that the activation level of the left and right abdominal internal oblique muscles  is 1\% when a person is walking normally. But when people walk with a single crutch, the activation levels of the left and right abdominal internal oblique muscles are 100\% and 50\%, respectively. It can be seen that when people walk with a single crutch, the left abdominal internal oblique muscles cause a lot of loads. This situation cause the asymmetric activation of the muscles. 
And the load on the left and right muscles is much higher than normal walking. When people walk with double crutches, the activation levels of the left and right abdominal internal oblique muscles are 81\% and 77\%, respectively. Although the muscles load is higher than when walking normally, it is more symmetrical than when people walk with a single crutch. The main roles  of these muscles are to maintain the stability of the trunk and to assist the external oblique muscles to help with the trunk rotation. The mechanism is similar to that of the external oblique muscles, but from the results, we can see that maintaining this stability consumes a lot of energy.

Abdominal external oblique muscles and abdominal internal oblique muscles are directly connected to the bones. When the muscles on both sides are in an asymmetric state for a long time, it is easy to cause the muscles on one side to be overdeveloped, causing the spine bend  to the side of the muscles and cause scoliosis.

The latissimus dorsi is a large, flat muscle on the back that stretches to the sides, behind the arm, and is partly covered by the trapezius on the back near the midline, shown in Figure~\ref{fig_latissimus dorsi}. The latissimus dorsi is responsible for extension, adduction, transverse extension also known as horizontal abduction, flexion from an extended position, and (medial) internal rotation of the shoulder joint. It also has a synergistic role in extension and lateral flexion of the lumbar spine. From Table~\ref{tab:Trunk_muscle_activation}, we can see that the left and right latissimus dorsi muscles activation levels are both 1\% when a person is walking normally. But when people walk with a single crutch, the left and right latissimus dorsi muscles activation levels are 15\% and 1\%. Although there is no additional burden on the right muscles, they are overloaded by the left muscles. It can be seen that when people walk with a single crutch, the left latissimus dorsi muscles cause a lot of burdens. This situation causes the asymmetric activation of the muscles. 
When people walk with a double crutches, the left and right latissimus dorsi muscles activation levels are both 1\%. The muscles symmetry of the double crutches is perfect, and the muscles load is minimal, which is similar to normal walking. 

Here we can see that when people walk with a crutch, the arm provides a part of the force, and latissimus dorsi muscles transfer this part of the force to the torso. Because the right foot is injured and the left hand uses a crutch. In this case the force on the left will increase. The right hand is not exerted extra load, so the activation of the right muscles is the same as normal walking. As this part of the muscles is directly connected to the spine, an asymmetrical state of the back muscles over a long period of time can bend the spine and thus create scoliosis.

\begin{figure}
    \centering
    \includegraphics[width=0.45\linewidth]{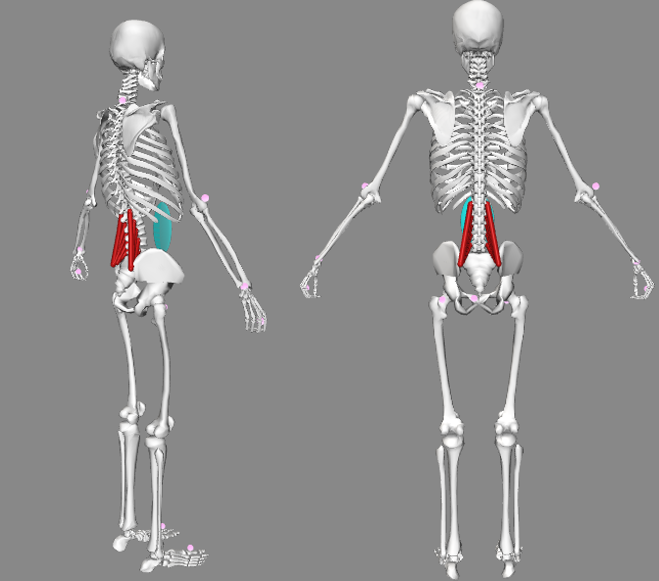}
    \caption{The longissimus dorsi (lumbar part) muscle group.}
    \label{fig_Longissimus}
\end{figure} 
The longissimus dorsi is lateral to the semispinalis, shown in Figure~\ref{fig_Longissimus}. It is the longest subdivision of the erector muscles of the spine and extends forward into the transverse processes of the posterior cervical vertebrae. From Table~\ref{tab:Trunk_muscle_activation}, we can see that the activation levels of the muscles of the left and right longissimus are both 1\% when a person walks normally. But when people walk with a single crutch, the activation levels of the left and right longissimus muscles are 80\% and 1\%. Although there is no additional load on the right muscles, they are overloaded by the left muscles. It can be seen that when people walk with a single crutch, the left longissimus muscles cause a lot of burdens. This situation causes the asymmetric of the muscle activation. 
When people walk with double crutches, the activation of the left and right longissimus muscles is 1\%. The muscles symmetry of the double crutches are perfect, and the muscles load are minimal, which are similar to normal walking.
\section{Discussions}
We simulated the process of walking with crutches, and obtained the levels of activation of the human trunk muscles during normal walking, using a single crutch and double crutches. By comparing the activation levels of these muscles, we can see that the use of crutches, especially a single crutch, can lead to a large difference in the activation level of the back muscles on the left and on the right sides. This long-term difference can lead to muscle degeneration and scoliosis. Compared to normal walking, the double crutches used in this experiment increased stress, but they are still acceptable. The symmetry of activation of the muscles on the left and right sides is better than using a single crutch for walking.
\section{Conclusions}
In this article, we found that people using double crutches for walking use less force than when they use single crutches and the activation of their muscles is more symmetrical.
However, although we have currently evaluated the use of single and double crutches, we only studied a small sequence for our analysis. Although the sequence we have chosen is very representative, more precise evaluations still require dynamic analysis. In addition, our  results come only from numerical simulations. 
In order to verify the accuracy of the simulations, real experiments will be conducted to verify these analyses using ART motion capture systems and force platforms. The information obtained will then be used in OpenSim to be compared with our simulations. A fatigue model will be used to identify the most stressed muscles and the impact of this fatigue in postural changes.
\bibliographystyle{splncs04}
\bibliography{hciRef}
\end{document}